\documentclass[11pt]{article}

\usepackage{acl}

\usepackage{times}
\usepackage{latexsym}
\usepackage[T1]{fontenc}
\usepackage[utf8]{inputenc}
\usepackage{microtype}
\usepackage{graphicx}
\usepackage{booktabs}
\usepackage{amsmath}
\usepackage{amssymb}
\usepackage{multirow}
\usepackage{xcolor}

\providecommand{\aclfinalcopy}{}
\aclfinalcopy

\title{Your Judge Inherits Your Prompt: In-Context Label Bias in an LLM-as-Judge Is Directional, Not Symmetric}

\author{
  Abdullah Ahmad\textsuperscript{1} \quad
  Ahmed J. O. Ahmed\textsuperscript{1} \quad
  Luca Citi\textsuperscript{2} \\[2pt]
  \textsuperscript{1}Independent Researcher \quad
  \textsuperscript{2}University of Essex \\[2pt]
  \texttt{\{abdullah.ahmad.ac, eng.ahmedjafer\}@gmail.com} \\
  \texttt{lciti@essex.ac.uk}
}

\begin{document}
\maketitle

\begin{abstract}
LLM-as-judge has become a standard way to score open-ended model outputs, yet the judge is itself a frozen language model steered by a few in-context examples, and such models are known to absorb the label balance of their demonstrations rather than treat them as neutral illustrations. We ask whether this label bias reaches an LLM judge used inside a multimodal (video question answering) evaluation pipeline, and what form it takes. Using a fully frozen generator--judge pipeline, we hold the generated answers fixed and vary only the Yes/No balance of the judge's few-shot examples across four demonstration sets. Each verdict is validated against a two-annotator human consensus on 200 examples per dataset. The judge's acceptance rate stays close to human consensus when its examples are balanced or skewed toward rejection, but jumps sharply (by about 30 points on NExT-QA and 20 on MSVD-QA, with a strongly one-sided error profile) when the examples are skewed toward acceptance; the opposite skew produces no matching drop. The bias is therefore one-directional within the tested range of label ratios. A separate five-draw analysis shows that results from a single example draw are unreliable, varying by up to 18 points, more than any difference between prompting conditions. We conclude that demonstration label bias in an LLM judge is a directional, inflation-only failure mode that can fabricate apparent benchmark gains but not losses, and that it is cheaply avoided by balancing the judge's examples and reporting results averaged over several draws.
\end{abstract}

\section{Introduction}

A growing share of NLP analysis and evaluation is mediated by a second language model: an ``LLM-as-judge'' that reads a model output, compares it against a reference or rubric, and emits a verdict \citep{zheng2023judging}. The convenience is real, but the judge is not a neutral instrument. It is a frozen LLM steered by an in-context prompt, and a substantial body of work on in-context learning (ICL) shows that frozen LLMs do not read their context neutrally. \citet{zhao2021calibrate} showed that GPT-3 classification is systematically biased toward the majority label of its demonstrations. \citet{min2022rethinking} showed that the label distribution of the demonstrations conditions the output more than their content correctness does. \citet{lu2022fantastically} showed that demonstration order alone can swing accuracy by up to 30 points.

If these findings transfer to the judge setting, the leniency of an LLM judge becomes partly a property of its prompt rather than of the outputs it scores, and any absolute score it produces is confounded with the label balance of its own demonstrations. Two gaps motivate this work. The first gap is empirical. The exposed population is large and unaudited: training-free multimodal VideoQA is a widely used pipeline family that is routinely scored by an in-context-prompted LLM judge, yet to our knowledge no such pipeline has ever audited the label balance of its judge's demonstrations. The second gap is conceptual: the \emph{shape} of the bias, whether it acts symmetrically or in one direction, has not been characterized, and it is the shape that determines whether the failure inflates scores, deflates them, or both. We are explicit that the judge itself is text-only: it compares generated text against reference text, and the multimodality lives on the generator side. The contribution is therefore not that multimodality changes the mechanism, but that it exposes a widely used, judge-scored population to a directional failure mode that has so far gone unaudited.

Across two video question answering benchmarks, we find that demonstration label bias does reach the judge, but asymmetrically. When the judge's few-shot examples are balanced or skewed toward rejection, its acceptance rate stays close to human consensus; only a strong skew toward acceptance moves it, and then sharply and one-sidedly, inflating measured accuracy while agreement with humans does not improve. The bias thus pushes reported accuracy up but never down, an inflation-only asymmetry that \citet{jain2025agreeableness} also observe behaviorally in another judge setting (high acceptance of valid outputs, weak rejection of invalid ones), and that we isolate causally here. Intuitively, the judge already leans toward acceptance, so a Yes-skew pushes with that lean while a No-skew must first cancel it (Section~\ref{sec:analysis}). A complementary analysis shows that reporting a prompting result from a single example draw is unreliable, because accuracy varies more across draws than across prompting conditions. Together these results identify a directional and cheaply mitigable failure mode in any evaluation pipeline built on a prompted LLM judge. Even if the mechanism behind the asymmetry is contested, the exemplar-variance result stands on its own: single-draw prompting reports are unreliable, with a +16.09-point one-shot lift from the best draw shrinking to a non-significant +5.88 five-draw mean ($t{=}1.93$, $p{=}0.126$).

\paragraph{Contributions.} (1) We show, in a controlled setting, that in-context label bias reaches an LLM judge and is \emph{directional}: a skew toward acceptance inflates the judge's verdicts, while an equal-and-opposite skew toward rejection does not deflate them. This is a rectified, inflation-only pattern. We isolate it with a fixed-generation intervention and interpret it as the ICL label-prior \citep{zhao2021calibrate} composing with the judge's known positivity disposition \citep{wang2023large}. (2) We quantify exemplar variance with a five-draw replication, showing that single-draw prompting effects are inflated relative to the population. (3) We contribute a dual-dataset human-validation protocol for LLM judges. (4) We provide cheap, mechanism-matched mitigations and release all prompts, exemplars, annotations, and code.\footnote{Code, prompts, exemplars, annotations, and per-cell $\kappa$ tables: \url{https://github.com/abuba8/LLM-Judge-Label-Bias}}

\section{Background and Related Work}

\paragraph{Label bias and exemplar sensitivity in ICL.} \citet{liu2022makes} and \citet{rubin2022learning} documented that exemplar choice materially affects GPT-3 accuracy. \citet{zhao2021calibrate} identified majority-label bias and proposed contextual calibration, a correction applied directly to the model's label-token priors; subsequent work refined this line, with \citet{fei2023mitigating} formalizing a typology of label biases and showing that prior calibration does not address all of them, and \citet{zhou2023batch} unifying these methods via batch calibration. Because the bias we characterize lives in the same label-token prior, this calibration family supplies its mechanism-matched fix. \citet{min2022rethinking} showed that the demonstrations' label distribution conditions output more than content correctness, and \citet{lu2022fantastically} showed order sensitivity of up to 30 points. Our work reproduces this family of results in a new locus, the demonstrations conditioning a judge's verdicts rather than a classifier's predictions, and adds a wrinkle specific to judges: the effect is directional. We interpret this through the judge's decision rule, an argmax over ``Yes''/``No'' logits, in which the demonstration prior acts as a bias term on the Yes logit that, given the judge's Yes-leaning operating point, saturates on the No side.

\paragraph{LLM-as-judge reliability.} \citet{zheng2023judging} showed that GPT-4 judgments correlate with human preferences; \citet{wang2023large} documented systematic leniency and positivity biases; and \citet{chiang2023llm} found that LLM judges give reliable relative rankings while their absolute calibration is dataset-dependent. A rapidly growing literature now catalogs judge biases \citep{gu2024survey}, most of them arising from the \emph{candidate responses}: position, verbosity, and self-preference are among the twelve types quantified by \citet{ye2025justice}. Judge verdicts are further destabilized by prompt- and rubric-side perturbations: paraphrasing or reformatting prompts can reverse model rankings \citep{mizrahi2024state}, rubric-order and score-ID changes shift scores \citep{scoringbias2025}, and judges are inconsistent even under fixed inputs \citep{stureborg2024inconsistent}. Our five-draw protocol is the demonstration-side analogue of these multi-prompt reliability calls. The bias we study is different in origin: it arises from the judge's own \emph{demonstrations} rather than from the responses being scored. Closest to our finding, \citet{jain2025agreeableness} observe an ``agreeableness'' profile in which judges accept valid outputs at high rates but fail to reject invalid ones (true-negative rates below 25\%), inflating apparent reliability under class imbalance, a behavioral signature of the same one-directional inflation we isolate mechanistically here. We do not merely restate that judges are lenient. We show that the ICL label-prior \emph{composes} with that leniency: a skew toward acceptance amplifies it into a large acceptance spike, while an equal skew toward rejection is absorbed and leaves the judge tracking human consensus. The two known biases are not independent, and their interaction is what makes judge label bias an inflation-only risk.

\paragraph{Training-free multimodal pipelines as the exposed population.} Training-free video QA steers frozen models with prompts: VideoAgent \citep{wang2024videoagent}, VidIL \citep{wang2022language}, and ViperGPT \citep{suris2023vipergpt}; Flamingo \citep{alayrac2022flamingo} noted exemplar sensitivity even in few-shot-trained models, and MM-REACT \citep{yang2023mmreact} showed that prompt structure determines output format. These works report single-exemplar numbers and typically score with an LLM judge; to our knowledge, none quantifies exemplar variance or audits judge-demonstration balance. Fine-tuned video QA systems \citep{jiang2020reasoning,gao2023mist,yu2024self,ye2023hitea,ko2023large} sidestep the judge via multiple-choice protocols but at a compute cost (a full fine-tune in our environment was projected at more than 140 GPU-hours per epoch), which is why the exposed training-free population keeps growing.

\section{Methodology}
\label{sec:method}

\begin{figure*}[t]
 \centering
 \includegraphics[width=\textwidth]{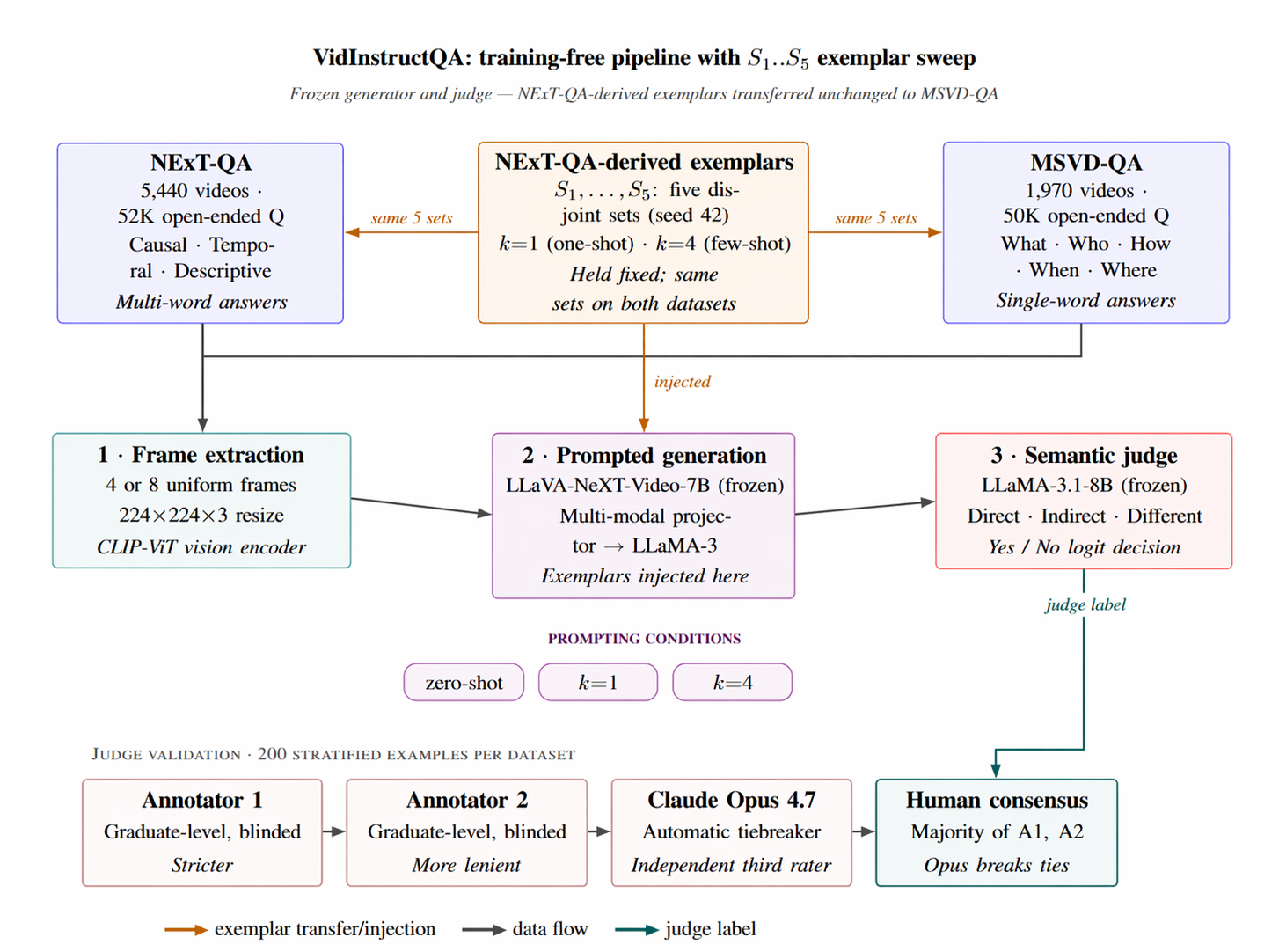}
 \caption{Generic structure: generator $\rightarrow$ judge $\rightarrow$ human validation, with one manipulated variable (the label balance of the judge's demonstrations); the figure shows our instantiation. The evaluation pipeline and judge-validation protocol. Three frozen stages: frame extraction, prompted generation (LLaVA-NeXT-Video-7B), and a semantic judge (LLaMA-3.1-8B) that emits a Yes/No decision from its output logits. The five exemplar sets S1--S5 are drawn once from NExT-QA train and transferred unchanged to MSVD-QA; they are injected at the prompted-generation stage and are therefore visible to \emph{both} the generator (as answer-format demonstrations) and the judge (as rubric-verdict demonstrations), the two conditioning channels analyzed in Section~\ref{sec:analysis}. Judge validation uses two blinded annotators with an automatic tiebreaker on disagreements; the resulting consensus is the reference used throughout.}
 \label{fig:pipeline}
\end{figure*}

Our goal is to isolate, and then characterize, the effect of the judge's demonstration labels on its verdicts. To do so we build a fully frozen pipeline (Figure~\ref{fig:pipeline}) in which nothing is trained and in which the only variable we deliberately manipulate is the Yes/No balance of the judge's in-context examples. This section describes the pipeline, the datasets and their question categories, the exemplar sweep that varies the label balance, and the human-validation protocol against which the judge is calibrated.

\subsection{Pipeline}

The pipeline has three frozen stages. \textbf{(1) Frame extraction} samples 4 or 8 uniform frames per video, resized to $224\times224\times3$. \textbf{(2) Prompted generation} uses LLaVA-NeXT-Video-7B-hf \citep{liu2024llavanext,zhang2024llavanextvideo}, entirely frozen; we use the April 2024 release, which predates the LLaVA-Video SFT corpus incorporating NExT-QA, so to our knowledge neither benchmark is in this model's training data. \textbf{(3) Semantic judging} uses a frozen LLaMA-3.1-8B \citep{meta2024llama} judge that scores the generated answer against the ground truth. The judge prompt defines three decision classes (Direct Match, Indirect Context, and Different Context), which collapse to a binary decision: the first two count as accept (Yes) and the third as reject (No). The emitted verdict is the higher-probability logit between the ``Yes'' and ``No'' tokens. This detail matters for interpretation, because majority-label bias is precisely a shift in this label-token prior, and it is therefore directly observable as a shift in the judge's acceptance rate. It is also why we study a binary judge: majority-label bias lives in the label-token prior \citep{zhao2021calibrate}, and a two-token argmax isolates it most cleanly.

\subsection{Datasets and Question Categories}

We pair two open-ended benchmarks with deliberately different answer distributions. NExT-QA \citep{xiao2021next} contains 5,440 videos (average 44s) and 52,044 questions whose ground-truth answers are often multi-word phrases. MSVD-QA \citep{xu2017video} contains 1,970 short clips (under 10s) and 50,505 questions whose answers are overwhelmingly single words. Pairing them lets us separate a generation-side format effect from the judge-side label effect. Both benchmarks organize questions into categories; Table~\ref{tab:categories} lists these categories together with their counts in the 200-example, type-stratified validation sample used for human validation. The multi-word versus single-word contrast, and the differing category mixes, are what allow us to distinguish effects that act through the generator from effects that act through the judge.

\begin{table}[t]
\centering
\small
\begin{tabular}{llcc}
\toprule
Dataset & Category & Corpus share & Val.\ $n$ \\
\midrule
\multirow{3}{*}{NExT-QA} & Causal & 48\% & 105 \\
 & Temporal & 29\% & 62 \\
 & Descriptive & 23\% & 33 \\
\midrule
\multirow{5}{*}{MSVD-QA} & What & n/a & 119 \\
 & Who & n/a & 74 \\
 & How & n/a & 6 \\
 & When & n/a & 1 \\
 & Where & n/a & 0 \\
\bottomrule
\end{tabular}
\caption{Question categories in each benchmark and their counts in the 200-example, type-stratified validation sample. NExT-QA shares are corpus-level; per-type corpus shares for MSVD-QA are not reported by the source, so we give validation counts only. The MSVD-QA sample under-represents How and When and contains no Where, a coverage caveat we carry forward.}
\label{tab:categories}
\end{table}

\subsection{Exemplar Sweep}
\label{sec:sweep}

Zero-shot prompting uses a system prompt plus the question. One-shot ($k{=}1$) adds a single worked exemplar (question, ground truth, generated answer, and a binary rubric verdict), and few-shot ($k{=}4$) adds four. By construction, an exemplar conditions the pipeline at two levels: at the generation level it shapes the format of the generator's answers, and at the judge level the full (generated answer, rubric verdict) pair is visible to the judge, so the verdicts act as in-context demonstrations of the rubric.

We draw five disjoint exemplar sets, S1 through S5 (seed 42), from 35,653 deduplicated NExT-QA training rows, for each of the one-shot and few-shot conditions, and transfer the same sets unchanged to MSVD-QA. Within each exemplar set, demonstration order is held fixed across all prompting conditions and datasets, so that any verdict change is attributable to label balance rather than to order effects \citep{lu2022fantastically}. The sets vary along two axes. The controlled, judge-relevant axis is the rubric-verdict balance: S2 is 1:3 Yes:No (Yes-fraction 0.25), S3 is 3:1 (0.75), and S4 and S5 are 2:2 (0.50). A second axis is category composition: S1, S2, S4, and S5 use a 2/1/1 Causal/Temporal/Descriptive split, whereas S3 uses 1/2/1. S1 predates the verdict-balance control (it was drawn for the initial one-shot pilot) and is excluded from the verdict-balance analysis. Because S3 differs on both axes, we do not rest the mechanism on S3's test accuracy alone; Section~\ref{sec:results} separates verdict balance from category composition using the fixed-generation control, the one-shot ($k{=}1$) replication, and a bounded-magnitude argument from the per-category accuracies. Each (dataset, frames, condition, set, split) cell is a single full-split run, for 80 full-split cells plus zero-shot baselines, on 3$\times$ NVIDIA A30 24GB GPUs at roughly 3 GPU-hours per split.

\subsection{Fixed-Generation Control and Human Validation}
\label{sec:validation}

The central control is that, on the 200-example validation samples, greedy generations are byte-identical across prompting conditions. Any verdict that changes with the prompt on these samples is therefore attributable to the judge's demonstrations alone; this is a fixed-input intervention that isolates the judge's decision computation from the generator. We state its scope precisely: the control separates \emph{judge-side} conditioning from \emph{generation-side} conditioning, because the generated text is held fixed, but it does not on its own separate the two axes that vary \emph{within} the judge-side demonstrations, verdict balance and category composition, which we disentangle in Section~\ref{sec:results} and Section~\ref{sec:analysis}.

For each dataset we sample 200 test examples stratified by question type (Table~\ref{tab:categories}). Two graduate-level annotators, blinded to the judge's decisions, independently label each (question, ground truth, generated answer) triple as accept or reject, and an independent LLM tiebreaker (Claude Opus 4.7) resolves disagreements under the same rubric. Inter-annotator agreement is 85.00\% on NExT-QA ($\kappa{=}0.693$) and 86.00\% on MSVD-QA ($\kappa{=}0.695$), substantial \citep{landis1977measurement} and nearly identical across datasets, so the resulting consensus is a reliable reference. The tiebreaker resolved 30/200 and 28/200 disagreements, siding with the more lenient annotator; because any leniency it shares would raise the consensus acceptance rate, its influence is conservative for the effect we report. Unparseable judge outputs are mapped to reject, and a drop-unparseable sensitivity analysis preserves all within-condition rankings.

\section{Results}
\label{sec:results}

\subsection{The Judge Inherits Its Demonstrations' Label Balance, in One Direction}

\textbf{Finding.} A skew toward acceptance lifts the judge's acceptance rate 20--30 points above human consensus with a one-sided error profile; an equal skew toward rejection leaves it on the consensus line. Few-shot set S3 produces an 82.22\% judge acceptance rate on NExT-QA and 82.67\% on MSVD-QA, roughly 26.8 and 15.6 points above the mean of the matched non-skewed few-shot sets (S2, S4, S5). Table~\ref{tab:table1} shows why this is inflation rather than improvement. Because each Yes-fraction level is represented by a single set per dataset, the rejection-side conclusion rests on two points (one Yes-skewed and one No-skewed set per dataset), so we ground it in the one-sided error profile, the depressed $\kappa$, and the one-shot replication below, rather than in the point estimates alone. On the fixed-generation validation samples, the judge's acceptance rate stays within 2.5 points of human consensus for every set at or below balance (Yes-fractions 0.25 and 0.50) on both datasets, and then jumps to 86.0\% on NExT-QA and 83.0\% on MSVD-QA, 29.5 and 19.5 points above consensus, at the single set skewed toward acceptance (S3). The skew toward rejection (S2) produces no matching downward departure: its acceptance rate is within 1.5 points of consensus. The two skewed sets sit at equal distance from balance (0.25 vs.\ 0.75), so the asymmetry holds at matched skew magnitude; whether a more extreme rejection skew would deflate the judge is untested, and our claim is scoped to this range.

\begin{table}[t]
\centering
\small
\setlength{\tabcolsep}{4pt}
\begin{tabular}{lccccccc}
\toprule
Set & Yes-fr. & J\% & C\% & Bias & FP & FN & FP:FN \\
\midrule
\multicolumn{8}{l}{\textit{NExT-QA}} \\
S2 & 0.25 & 58.0 & 56.5 & +1.5 & 33 & 30 & 1.1 \\
S4 & 0.50 & 55.5 & 56.5 & $-$1.0 & 40 & 42 & 1.0 \\
S5 & 0.50 & 55.0 & 56.5 & $-$1.5 & 33 & 36 & 0.9 \\
\textbf{S3} & \textbf{0.75} & \textbf{86.0} & 56.5 & \textbf{+29.5} & \textbf{66} & \textbf{7} & \textbf{9.4} \\
\multicolumn{8}{l}{\textit{MSVD-QA}} \\
S2 & 0.25 & 64.0 & 63.5 & +0.5 & 31 & 30 & 1.0 \\
S4 & 0.50 & 61.5 & 63.5 & $-$2.0 & 35 & 39 & 0.9 \\
S5 & 0.50 & 66.0 & 63.5 & +2.5 & 32 & 27 & 1.2 \\
\textbf{S3} & \textbf{0.75} & \textbf{83.0} & 63.5 & \textbf{+19.5} & \textbf{52} & \textbf{13} & \textbf{4.0} \\
\bottomrule
\end{tabular}
\caption{Few-shot judge versus human consensus on the fixed-generation 200-example validation samples. J\% is the judge acceptance rate, C\% the human-consensus acceptance rate, and Bias $=$ J $-$ C. FP and FN are raw false-accept and false-reject counts (out of 200), and FP:FN their ratio. The acceptance rate stays within $\pm$2.5 points of consensus for every set at or below balance, and departs, upward only and with a strongly one-sided error profile (66 vs.\ 7 at S3 on NExT-QA), at the single Yes-skewed set.}
\label{tab:table1}
\end{table}

\begin{figure*}[t]
 \centering
 \includegraphics[width=\textwidth]{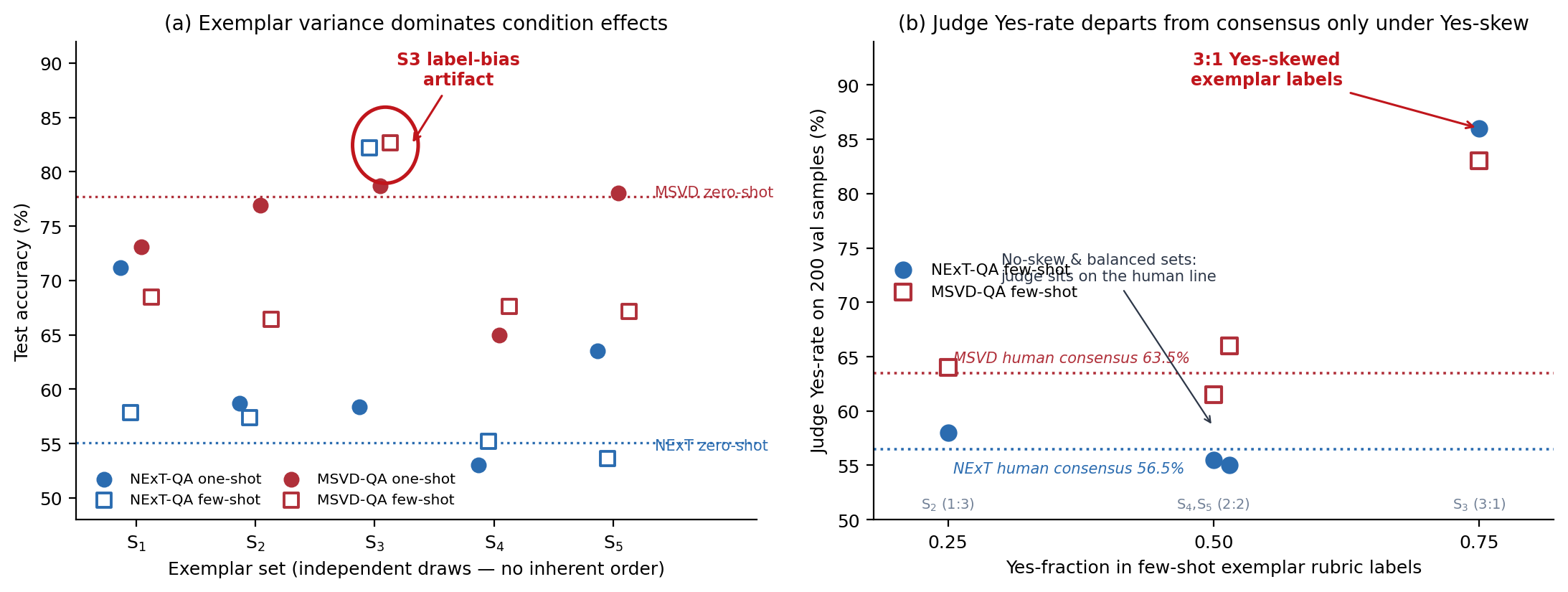}
 \caption{Main results. \textbf{(a)} Per-set test accuracy at 4 frames. The five exemplar sets are independent draws with no inherent order, so they are shown as unconnected points; the spread of the one-shot draws exceeds every condition-level gap, and the sole few-shot outlier is the S3 label-bias artifact. \textbf{(b)} Judge acceptance rate on the fixed-generation validation samples against the Yes-fraction of the few-shot exemplar labels, with human-consensus reference lines. The judge sits on the human-consensus line for the rejection-skewed (0.25) and balanced (0.50) sets and departs, upward only, at the single acceptance-skewed set (0.75). Points are deliberately not connected: the departure is a single-condition, one-directional jump, not a monotone trend.}
 \label{fig:results}
\end{figure*}

The error profile confirms the direction. Only S3 develops a one-sided profile, with false accepts far exceeding false rejects (66 vs.\ 7 on NExT-QA, a ratio of 9.4, and 52 vs.\ 13 on MSVD-QA, a ratio of 4.0), against a near-even split for every other set (e.g.\ 33 vs.\ 30 for the rejection-skewed S2 on NExT-QA). Agreement with humans does not improve at S3: its $\kappa$ is 0.195 (NExT-QA) and 0.209 (MSVD-QA), no better than the balanced sets. Figure~\ref{fig:results}b visualizes the pattern. Section~\ref{sec:analysis} explains why the response is rectified rather than symmetric.

We also report the correlation between demonstration Yes-fraction and judge acceptance rate: $r=0.76$ (NExT-QA) and $r=0.79$ (MSVD-QA). This four-point statistic is carried entirely by S3, however: removing it reverses the correlation to $r=-0.99$ (NExT-QA) and collapses it to $r=-0.06$ (MSVD-QA). We report the reversal because it is the substantive result: the relationship is a single-condition jump, not a smooth trend, and we therefore do not describe the effect as a monotone dose--response.

A partially independent signal at one-shot ($k{=}1$) corroborates the direction. Grouping the sets by the polarity of their single demonstration verdict, the Yes-labelled sets outscore the No-labelled sets by +5.08 points on NExT-QA (60.98 vs.\ 55.90) and +7.41 on MSVD-QA (78.38 vs.\ 70.97). This cut is orthogonal to S3's category confound: the Yes-labelled group includes S5, which uses the standard 2/1/1 category split rather than S3's 1/2/1, and S5 alone outscores both No-labelled sets on both datasets (63.56 vs.\ 58.74 and 53.06; 78.05 vs.\ 76.95 and 64.99). The acceptance-direction effect therefore appears in a set that does not share S3's category mix, pointing to verdict polarity rather than category composition. We are explicit that this uses a single demonstration ($k{=}1$) rather than the four-shot count of the headline effect, and that with two sets per polarity it is descriptive; it is thus strong corroboration of the direction, not a clean replication of the four-shot magnitude.

Throughout, we read these Yes-rate departures as measurements of judge leniency, not as accuracy. Judge--human $\kappa$ is only fair (0.17--0.36 across the grid, well below the 0.69 inter-annotator agreement), so the absolute accuracies in Table~\ref{tab:table2} are best interpreted as relative comparisons across conditions rather than as calibrated scores.

\subsection{Exemplar Variance Exceeds Condition Effects}

\textbf{Finding.} The spread across five exemplar draws exceeds every mean difference between prompting conditions, so a single-draw prompting result is not interpretable on its own. Across the five draws, one-shot NExT-QA test accuracy at 4 frames spans 18.14 points (from 53.06 to 71.20; mean 60.99$\pm$6.81), and MSVD-QA spans 13.73 points (mean 74.37$\pm$5.67). Excluding the S3 artifact, the few-shot sets are nearly identical, collapsing to a 4.24-point range on NExT-QA (56.03$\pm$1.97) and 2.08 points on MSVD-QA (67.42$\pm$0.87); Figure~\ref{fig:results}a shows this directly. This variance exceeds every mean difference between prompting conditions: on NExT-QA the mean one-shot advantage over zero-shot is +5.88 with a within-condition standard deviation of 6.81. The single-draw number a conventional protocol would report (+16.09 on NExT-QA, from set S1) shrinks to a five-draw mean of +5.88 that is not statistically distinguishable from zero at $n{=}5$ ($t{=}1.93$, $p{=}0.126$). Table~\ref{tab:table2} reports all test cells.

\begin{table*}[t]
\centering
\small
\begin{tabular}{llccccccccc}
\toprule
Dataset & Cond. & Frames & Zero & S1 & S2 & S4 & S5 & Mean* & Std* & S3 (artifact) \\
\midrule
NExT-QA & one-shot & 4 & 55.11 & 71.20 & 58.74 & 53.06 & 63.56 & 60.99 & 6.81 & 58.40 \\
NExT-QA & one-shot & 8 & 54.72 & 71.67 & 58.57 & 52.31 & 63.47 & 60.98 & 7.18 & 58.89 \\
NExT-QA & few-shot & 4 & 55.11 & 57.86 & 57.39 & 55.24 & 53.62 & 56.03 & 1.97 & \textbf{82.22} \\
NExT-QA & few-shot & 8 & 54.72 & 58.12 & 57.36 & 55.37 & 54.33 & 56.30 & 1.75 & \textbf{82.92} \\
MSVD-QA & one-shot & 4 & 77.72 & 73.12 & 76.95 & 64.99 & 78.05 & 74.37 & 5.67 & 78.72 \\
MSVD-QA & one-shot & 8 & 76.73 & 72.36 & 76.01 & 64.12 & 77.04 & 73.41 & 5.58 & 77.54 \\
MSVD-QA & few-shot & 4 & 77.72 & 68.50 & 66.42 & 67.61 & 67.15 & 67.42 & 0.87 & \textbf{82.67} \\
MSVD-QA & few-shot & 8 & 76.73 & 68.15 & 66.03 & 66.01 & 67.44 & 66.91 & 1.07 & \textbf{83.36} \\
\bottomrule
\end{tabular}
\caption{Judge-scored test accuracy, i.e.\ the judge's per-cell verdict (acceptance) rate (\%); these are the judge's verdict rates, not accuracies against human ground truth. The flagged \textbf{S3 few-shot} cells (bold) are the label-bias artifact and must not be read as system accuracy. One-shot Mean/Std are over all five sets; few-shot Mean/Std exclude the S3 artifact, which is listed separately. Validation-split numbers (within $\leq$2 points of test in all cells) ship with the code.}
\label{tab:table2}
\end{table*}

\subsection{A Smaller Directional Transfer Effect}

\textbf{Finding.} The one-shot effect is directional across datasets (positive on NExT-QA, negative on MSVD-QA), but its five-draw magnitude is roughly a third of the single-draw estimate. At the mean-across-sets level, a directional cross-dataset pattern replicates: the one-shot mean exceeds zero-shot on NExT-QA in all four (split $\times$ frames) cells (+5.88 to +7.38) and falls below zero-shot on MSVD-QA in all four cells ($-$2.40 to $-$3.68). Because the validation and test splits agree to within two points, we treat this as one directional observation rather than four independent ones. A single-draw view suggests a $+16/-5$ effect, whereas the five-set population effect is roughly $+6/-3$, with the positive side not significant at $n{=}5$.

\section{Analysis}
\label{sec:analysis}

\paragraph{The bias is directional, and why that matters.} The results describe a judge whose acceptance rate is well calibrated to human consensus until its demonstrations are skewed toward acceptance, at which point it decouples upward and only upward, a rectified response. We interpret this as the ICL label-prior of \citet{zhao2021calibrate} composed with the judge's pre-existing positivity disposition \citep{wang2023large}. Reading the verdict as an argmax over the ``Yes''/``No'' logits, the demonstration prior adds a bias to the Yes logit. The judge already rests slightly above human consensus, so a skew toward acceptance amplifies a disposition the model is already inclined toward, flipping a large mass of borderline rejects to accepts and producing the false-accept explosion. A skew toward rejection, by contrast, must first overcome that resting lean, and largely fails, so the response saturates on the rejection side. This composition, rather than either bias alone, is what makes the effect one-directional. We present this saturating-prior account as an interpretation consistent with every diagnostic we observe (the one-sided error profile, the depressed $\kappa$, and the $k{=}1$ direction), not as a direct measurement: confirming it would require logging the judge's Yes/No logit margin per set, which we did not record and leave to future work.

\paragraph{Two channels of conditioning.} The same exemplars condition the pipeline in two separable ways, and the analysis lets us tell them apart. Judge-side conditioning shapes the decision rule and is isolated by the fixed-generation control, where the input is held constant and only the demonstrations vary. Generation-side conditioning shapes the format of the generated answers and surfaces in the smaller cross-dataset transfer effect: multi-word exemplars push the generator toward longer outputs, which helps against NExT-QA's multi-word references and hurts against MSVD-QA's single-word labels. The exemplar variance documented in the results is the residual that both channels jointly produce, and it is large enough that a single draw sits well inside the noise of the exemplar distribution.

\paragraph{Strengths and weaknesses of the approach.} The main strength of the design is that the fixed-generation control gives a clean causal separation: because the generated text is byte-identical across conditions, any change in the verdict is attributable to the judge's demonstrations, and no appeal to generation quality can explain the acceptance spike. The main weakness is that the sharpest evidence comes from a single skewed condition (S3), which differs from the other sets on category composition as well as verdict balance. Three arguments, all from data already in the paper, localize the effect to verdict balance rather than category. First, the fixed-generation control holds the generated text fixed, so a category effect operating through generation quality cannot explain the acceptance spike. Second, judge accuracy varies only modestly across the three NExT-QA question categories, a range of about 9 points (Causal 56.29 to Descriptive 65.64; Table~\ref{tab:table3}), so a shift in exemplar category mix cannot plausibly produce a 20--30-point spike. Third, the one-shot direction holds for S5, a standard-category set, cutting across the category axis. A category-matched skewed set would still make the case airtight and remains the cleanest follow-up. A related weakness is that agreement with humans, measured by $\kappa$, is only fair across the grid, so the absolute accuracies should be read as relative comparisons rather than calibrated scores.

\paragraph{Implications for mitigation.} We set out to test whether demonstration label bias reaches an LLM judge and, if so, what form it takes. The answer is that it does, and that its form is directional: the judge can be prompted into inflation but not, in this operating range, into deflation. This sharpens the generic caution that ``LLM judges are lenient'' into a specific, controllable, and one-sided mechanism, and it implies that mitigation can be targeted: balancing the demonstration verdicts removes a mechanism that can only ever inflate, and label-prior calibration \citep{zhao2021calibrate, fei2023mitigating, zhou2023batch} is the mechanism-matched fix, since the bias lives in the label-token prior. Two clarifications. First, balancing is not a vacuous recommendation: demonstrations are usually chosen as illustrative ``good'' cases or drawn from a strong system's outputs, both predominantly accept-labelled, so the naive protocol lands in the inflating regime, and no pipeline we know of audits this. Second, balancing fixes score inflation, not judge reliability: item-level agreement remains only fair ($\kappa$ 0.17--0.36) even with balanced demonstrations, so human validation on the target distribution is still required. Applying these label-prior calibration methods as an empirical mitigation, and measuring how much of the acceptance spike they remove, is the natural next step; because our bias occupies the same label-token prior they correct, we expect it to be directly addressable, and we leave this empirical test to future work. That the failure is inflation-only is corroborated from a different angle by \citet{jain2025agreeableness}, who report the same high-acceptance, low-rejection asymmetry behaviorally in a separate domain; our fixed-generation control adds the causal attribution, tracing that asymmetry to the demonstration label balance rather than to generation quality or task difficulty.

\section{Conclusion}

We asked whether an LLM judge inherits the label balance of its in-context demonstrations, and found that it does, but in only one direction within the tested range of label ratios. Holding the generated answers fixed and varying only the Yes/No balance of the judge's few-shot examples, we showed that a skew toward acceptance lifts the judge's acceptance rate far above human consensus, one-sidedly and without improving agreement, inflating matched benchmark accuracy by 26.8 points on NExT-QA (15.6 on MSVD-QA) and departing from human consensus by 29.5 (19.5), whereas an equal skew toward rejection leaves the judge tracking human consensus. A complementary analysis showed that single-example reporting is unreliable, with variance across draws exceeding every condition-level difference. Demonstration label bias in an LLM judge is thus a directional, inflation-only failure mode with two cheap remedies, balancing the judge's demonstrations and averaging over several draws. Under such a judge, a reported benchmark accuracy can be pushed up but not down. Balancing removes the inflation but not the judge's item-level unreliability, so it complements rather than replaces human validation.

\paragraph{Limitations.} The asymmetry rests on a single rejection-skewed set per dataset, corroborated by its balanced errors and highest agreement but not yet traced as a full curve. The acceptance-skewed set also differs in category composition; the fixed-generation control and the per-category bound (Section~\ref{sec:analysis}) make this an unlikely alternative explanation, though a category-matched set would settle it definitively. The proposed saturating-prior mechanism is likewise an interpretation rather than a direct measurement. Our results concern a binary accept/reject judge; whether the pattern holds for Likert-scale or pairwise judges is untested. The results are further bounded by a single generator--judge pair, single runs per cell, limited coverage of some MSVD-QA question types, and a human reference whose hardest cases are settled by an automatic tiebreaker, though its direction is conservative for our finding.

\paragraph{Future work and recommendations.} A graded Yes-fraction sweep with a category-matched skewed set, paired with per-set logit-margin logging, would convert our single-condition result into a measured, mechanistically confirmed dose--response, and additional backbones would test its generality. In the meantime, any pipeline scoring with a prompted LLM judge should balance its demonstration labels (or apply contextual calibration), report a mean and standard deviation over at least four exemplar draws, include a matched zero-shot baseline, and validate against humans on each target distribution.

\bibliography{anthology}

\appendix

\section{Per-Category Breakdown}
\label{sec:appendixA}

Table~\ref{tab:table3} breaks down one-shot NExT-QA accuracy by question category, contrasting the single-draw S1 view with the five-draw mean. Two points support the main text. First, the single-draw figures overstate every category, most sharply for Temporal, where S1 (73.77) sits 8.3 points above the five-draw mean (65.46), mirroring the exemplar-variance result of Section~\ref{sec:results} at the category level. Second, cross-category differences are small: the five-draw means span only about 9 points (Causal 56.29 to Descriptive 65.64), which bounds how much S3's altered exemplar category mix could move accuracy and underwrites the category-confound argument in Section~\ref{sec:analysis}.

\begin{table}[h]
\centering
\small
\begin{tabular}{lcccc}
\toprule
Category & Zero & S1 & Mean & Std \\
\midrule
Causal & 53.93 & 71.11 & 56.29 & 9.96 \\
Temporal & 58.05 & 73.77 & 65.46 & 5.79 \\
Descriptive & 53.87 & 67.98 & 65.64 & 3.40 \\
Overall & 55.11 & 71.20 & 60.99 & 6.81 \\
\bottomrule
\end{tabular}
\caption{Per-category NExT-QA one-shot accuracy (\%) at 4 frames, test split. Cross-exemplar variance within a category (for example, Causal 56.29$\pm$9.96) is large relative to cross-category differences; the single-draw Temporal figure (73.77) sits 8.3 points above the five-draw mean, mirroring the exemplar-variance result at the category level.}
\label{tab:table3}
\end{table}

\section{Reproducibility Checklist Details}

Five disjoint sets per condition, seed 42, drawn from 35,653 deduplicated NExT-QA training rows; all 25 exemplars released verbatim. Compute: 3$\times$ NVIDIA A30 24GB, roughly 3 GPU-hours per split. Greedy decoding throughout, with validation generations byte-identical across conditions. Unparseable judge outputs map to reject; the drop-unparseable sensitivity analysis preserves within-condition rankings on both datasets. Frame count is inert: every 4-frame and 8-frame accuracy pair in Table~\ref{tab:table2} agrees to within about one point, ruling out frame sampling as a factor in any comparison. Full per-cell $\kappa$ tables and the with-S3 and pure-human recuts ship with the released code. Per-set Yes/No logit margins were not recorded in these runs; logging them is the direct probe of the mechanism proposed in Section~\ref{sec:analysis}.

\end{document}